\documentclass[letterpaper, 10pt, conference, twoside]{ieeeconf}

\IEEEoverridecommandlockouts

\usepackage{graphics} 
\usepackage{graphicx,color}
\usepackage{eso-pic}
\usepackage{epsfig} 
\usepackage{times} 
\usepackage{amsmath} 
\usepackage{amssymb}  

\usepackage{xcolor} 
\usepackage{color,soul}
\usepackage{breakurl} 
\usepackage[hidelinks, breaklinks, bookmarks]{hyperref}
\usepackage{hyperref, bookmark}

\usepackage{microtype}
\usepackage{booktabs} 
\usepackage{cancel}
\usepackage{float}
\usepackage{stfloats}	
\usepackage{todonotes}
\usepackage{arydshln}
\usepackage{makecell, booktabs, caption}
\usepackage{multirow}
\usepackage[backend=biber,
    style=numeric,
    sorting=none,
    firstinits=true,
    autocite=superscript,
    indexing=cite,
    pagetracker=spread,
    hyperref=true,
    maxcitenames=1,
    maxbibnames=3,
    doi=false,
    isbn=false,
    url=false,
    eprint=true,
    natbib]{biblatex}
\renewbibmacro{in:}{} 

\usepackage{microtype}
\usepackage[subtle,tracking=normal]{savetrees}

\usepackage{tikz}
\newcommand*\circled[1]{\tikz[baseline=(char.base)]{
        \node[shape=circle,draw,inner sep=0.5pt] (char) {\scriptsize #1};}}

\title{\LARGE \bf
    Words2Contact: Identifying Support Contacts from Verbal Instructions Using
    Foundation Models
}

\author{Dionis Totsila, Quentin Rouxel, Jean-Baptiste Mouret, Serena Ivaldi\\%
\thanks{This research was supported by the CPER CyberEntreprises, the
    Creativ'Lab platform of Inria/LORIA, the EU Horizon project euROBIN (GA
    n.101070596), the France 2030 program through the PEPR O2R projects  AS3 and
    PI3 (ANR-22-EXOD-007, ANR-22-EXOD-004).}
\thanks{All authors are affiliated with Inria, Universit\'e de Lorraine,
CNRS, Loria, F-54000. Contacts: {\tt\footnotesize
firstname.lastname@inria.fr}}%
}

\begin{document}

\AddToShipoutPicture*{%
    \AtPageLowerLeft{%
        \parbox[t][\paperheight][t]{\paperwidth}{%
            \vspace*{-1.5cm}
            \begin{center}

            \textbf{Full Reference:} D. Totsila et al. ``Words2Contact: Identifying Support Contacts from Verbal Instructions Using Foundation Models''. \\ IEEE-RAS Humanoids 2024.
            \href{https://ieeexplore.ieee.org/abstract/document/10769902}{https://ieeexplore.ieee.org/abstract/document/10769902}
        \end{center}
        }
    }
}
\maketitle
\thispagestyle{empty}
\pagestyle{empty}

\begin{abstract}
    This paper presents Words2Contact, a language-guided multi-contact
    placement pipeline leveraging large language models and vision language models.
    Our method is a key component for language-assisted teleoperation and
    human-robot cooperation, where human operators can instruct the robots where to
    place their support contacts before whole-body reaching or manipulation using
    natural language. Words2Contact transforms the verbal instructions of a human
    operator into contact placement predictions; it also deals with iterative
    corrections, until the human is satisfied with the contact location identified
    in the robot's field of view.
    We benchmark state-of-the-art LLMs and VLMs for size and performance in
    contact prediction. We demonstrate the effectiveness of the iterative
    correction process, showing that users, even naive, quickly learn how to
    instruct the system to obtain accurate locations.
    Finally, we validate Words2Contact in real-world experiments with the Talos
    humanoid robot, instructed by human operators to place support contacts on
    different locations and surfaces to avoid falling when reaching for distant
    objects.
\end{abstract}


\section{Introduction}

\begin{figure}[!th]
    \centering
    \includegraphics[width=\columnwidth]{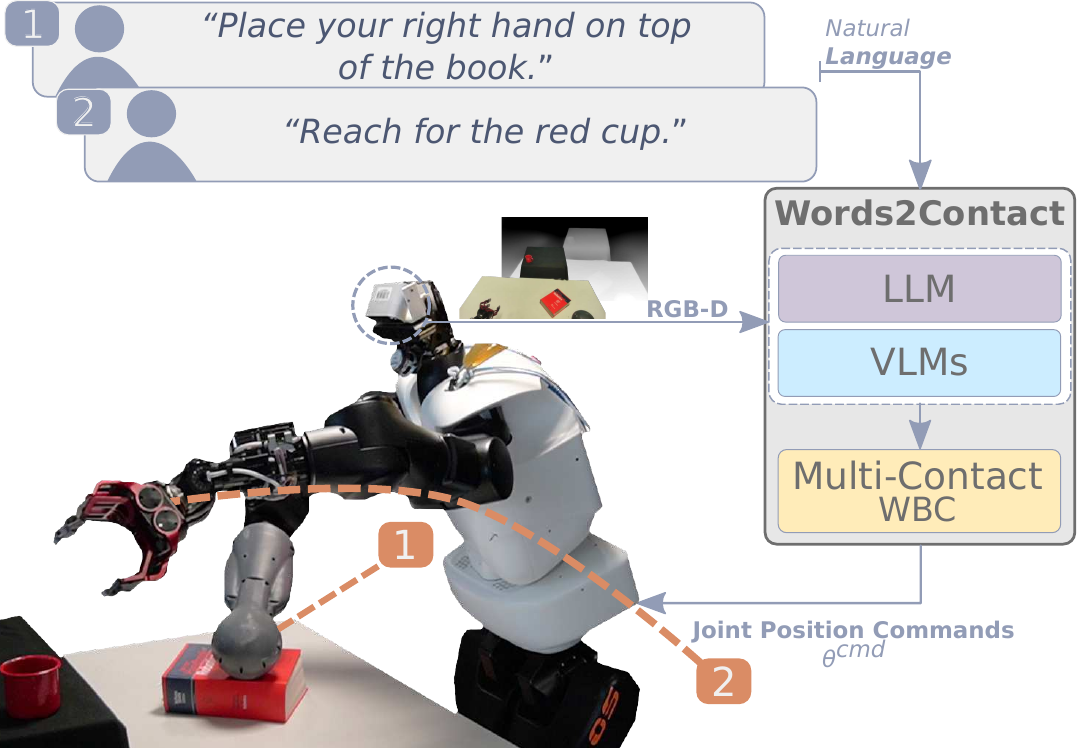}
    \vspace{-1em}
    \caption{Talos executes the user's verbal instructions to (1) lean on a
        book and (2) reach for an inaccessible cup, yet in its field of view, using our
        \textbf{Words2Contact} pipeline.}
    \label{fig:concept-figure}
\end{figure}

Humanoid robots can use various body parts to create support contacts to help
balance when reaching for difficult positions. For example, they can use their
right hand as a support on a table, bend forward and reach a cup that would
otherwise be out of reach (Fig.~\ref{fig:concept-figure}); or they can lean on
the counter with their left hand to reach for a dish in the bottom rack of a
dishwasher to prevent falling.
Solving these tasks autonomously is usually done with multi-contact whole-body
planners and controllers \cite{bouyarmane2018,padois2017whole}.


Recent advances in whole-body control using quadratic programming have shown
that both torque-controlled robots \cite{henze2016passivity} and
position-controlled robots with force/torque sensors \cite{rouxel2024multi} can
effectively utilize additional contact points to increase their manipulability
and improve their balance, but these control methods require the prior
knowledge of the contact locations. This information is usually the output of a
contact planning algorithm, where typically a planner decides a sequence of
contact locations that enable the robot to solve its task (e.g., walking,
manipulating a complex object) \cite{Kumagai2023}. Contacts computation usually
relies on visual or 3D perception and environment models to look first for
suitable contact surfaces, before deciding whether they are kinematically
feasible for the robot. For example, it is common to look for flat areas to
place the footsteps in humanoid walking~\cite{calvert2022fast}.

Unfortunately, selecting contacts, especially when applying forces, often
requires an understanding of the world that can hardly be modeled. Some
surfaces might be flat, but too fragile for support, like a glass window. Other
surfaces might be off-limits for safety reasons, like the wing of an aircraft,
or they might be slippery, dirty, or unstable. Overall, in many real-world
situations, the choice of support contacts is likely to require human expertise
at some point to be deployed outside of a laboratory.
Giving the power to human experts to guide the robot and choose the contact
locations for them is therefore a very desirable feature.

Human guidance in contact selection is ideal for teleoperated robots in remote
maintenance or hazardous scenarios and for collaborative robots cooperating and
working side-by-side with humans.
For example, a remote operator could instruct the robot to place one
end-effector on a wall to lift one foot, and a factory worker could instruct
the robot to reach a handle with one end-effector and take a fallen tool with
the other one. In these situations, language-based instructions provide a
natural communication channel and free the hands of the operator, nor do
constrain the human worker to use computer interfaces to instruct the robot on
what to do.

Giving instructions in natural language has long been a dream of the robotics
community~\cite{antoniol1993robust,tellex2020robots,mandery2016using}. For
years, this goal eluded researchers due to two main challenges: (1)
understanding what a sentence means requires a good intuition of the context
and the implicit knowledge, that is, some ``common sense'' (2); there are
countless ways of expressing the same instruction, which prevents the use of
simple keywords. To give an illustration, some people might refer to the
``Handbook of Robotics'' of Fig.~\ref{fig:concept-figure} as ``the big red
book'', ``the book'',``the book next to your right hand'', ``the red thing'',
``the big thing in front of you'', and so on.

Large Language Models (LLMs)~\cite{zhao2023surveylargelanguagemodels} might be
on the verge of solving these challenges for robotics
\cite{wang2024largelanguagemodelsrobotics}, providing a way to give natural and
general instructions to robots. Trained on billions of human-written texts,
LLMs exhibit a form of ``common sense'' that allows them to interpret
instructions with their most likely meaning. They are also, perhaps
surprisingly, highly versatile, as they are capable of handling instructions or
situations not anticipated by the robot designers. Additionally, LLMs naturally
process the many ways humans express similar concepts, as they are represented
by similar ``embeddings''. Visual Language Models (VLMs) are equally appealing,
as they can link text to images and vice-versa.

In this paper, we harness the power of Foundation Models (LLMs and VLMs) to
instruct a humanoid robot about desired contact locations for increased support
in whole-body reaching, an essential skill for solving several downstream
tasks.

The robotics community has been working intensely on integrating LLMs with
robots since the first demonstrations of ChatGPT (2022). The key challenge is
connecting perception, which is continuous, structured, and high-dimensional,
to language, which is linear and loosely structured, and then to actions, which
are also continuous and depend on the specific robot. While many approaches
have been proposed (see Sec.~\ref{sub-section:sota}), there is currently no
consensus on how to establish this link in the general case.

We present the following key contributions:
\begin{itemize}
    \item \textbf{Words2Contact}: a novel pipeline integrating LLMs and VLMs
          with a multi-contact whole-body controller to identify support contacts from
          verbal instructions.
    \item A benchmark of state-of-the-art LLMs and VLMs for contact prediction.
    \item A pilot study showing that users quickly learn to use our system to
          identify accurate contact locations.
    \item Validation of our system on a real Humanoid Robot.
\end{itemize}
To the best of our knowledge, this paper is the first to address support
contact identification from verbal instructions using Foundation Models and
demonstrate it with a humanoid robot.

\section{Related work}

\subsection{Multi-Contact and Whole-Body Control}
Multi-contact tasks present both theoretical and practical challenges that must
be addressed carefully. In these scenarios, humanoid robots exhibit
redundancies in their whole-body kinematic and contact force
distribution~\cite{seiko}, for example a robot can place its hand on a table in
multiple positions, with various combinations of force distribution for each
configuration. For humanoid robots, maintaining balance and ensuring the
feasibility of commanded motions are crucial for safety, and require
consideration of the system's physical constraints. Robustness is essential for
real hardware deployment, necessitating real-time control of posture and
contact forces.

Low-level control in humanoid robots is often achieved via Quadratic
Programming (QP) for whole-body optimization \cite{escande2014hierarchical},
allowing fast computation while accounting for constraints and allowing direct
control of contact forces on torque-controlled robots
\cite{henze2016passivity}.

However, such robots are sensitive to model errors, affecting robustness. In
this work, we conducted hardware experiments on the full-size humanoid robot
Talos \cite{stasse2017talos} controlled in position. In this context, contact
forces can be indirectly tracked through admittance schemes
\cite{caron2019stair}.

Our proposed contact selection method is agnostic to how multi-contact motion
is achieved. We utilized our SEIKO (Sequential Equilibrium Inverse Kinematic
Optimization) framework \cite{rouxel2024multi} for the multi-contact
experiments. SEIKO uses a Sequential QP to formulate a whole-body admittance
controller, explicitly modeling joint flexibility to indirectly regulate
contact forces on position-controlled robots.

\subsection{Robotics and Language Models}
\label{sub-section:sota}

Prior to the introduction of LLMs, numerous approaches in both language
comprehension and generation were explored in robotics~\cite{tellex2020robots,
    cangelosi2010integration}. However, these early methods
were limited due to their reliance on rigid, rule-based systems and predefined
vocabularies~\cite{mandery2016using}.

Thanks to their training on a very large dataset, LLMs can answer to a very
large set of natural language queries without having been trained on any
specific domain.

In particular they can provide high-level plans with some ``common sense'' by
inferring many pieces of context, thus bypassing most of the ``frame problem''
\cite{dennett1990cognitive}. In robotics, by using well-designed ``prompts''
that explain the problem to be solved in natural language and the kind of
expected output, LLMs were used to find a sequence of pre-learned behaviors
\cite{song2023llm,ren2023robotsaskhelpuncertainty}, generate Python code to be
executed by the robot
\cite{singh2022progpromptgeneratingsituatedrobot,liang2023codeaspolicies}, or
cost functions for a model-based controllers \cite{moweroptimal}. For example,
``Inner Monologue'' \cite{huang2023reasoninglargelanguagemodels} uses the
ability of LLMs to generate task plans and explores embodied reasoning through
self-dialogue. ``Code-as-Policies'' \cite{liang2023codeaspolicies} uses the
code generation abilities of LLMs to inform robotic policies directly from
natural language descriptions without the need for further training.

In some cases where the desired structure of the output is in a form that LLMs
are not inherently able to generate, or if the nature of the problem requires
more complex responses, additional fine-tuning may be useful
\cite{wu2023embodiedtaskplanninglarge,jin2023alphablockembodiedfinetuningvisionlanguage,kant2022housekeeptidyingvirtualhouseholds}.
For example, in ``BTGenBot'', behavior trees are generated through LLMs that
have been fine-tuned on specialized datasets. The key takeaway is that a
well-structured output is beneficial to transition from non-structured
high-level instructions to low-level control commands. The drawback of
fine-tuning is that it requires large amounts of data, which is time-consuming
and resource-intensive to collect, and may introduce biases based on how the
data are collected or generated \cite{yu2023largelanguagemodelattributed}.

Even though the generated plans are often successful, the ability to use
language-based corrections to fix the generated plans generated with minor
adjustments during task execution can be very useful. For example, Sharma et
al. \cite{sharma2022correctingrobotplansnatural} present a model that
integrates natural language and visual feedback to adjust robot planning costs
in real-time, enabling more dynamic and responsive adaptation to new tasks.
LILAC \cite{cui2023onlinelanguagecorrections} proposes a shared autonomy
paradigm that updates the control space in response to continuous user
corrections. DROC \cite{zha2024distilling} further advances this paradigm by
enabling LLM-based robot policies to respond to, remember, and retrieve
feedback efficiently, significantly improving adaptability to natural language
instructions. Overall, a correction mechanism that understands general and
abstract corrections, such as \textit{``a bit to the right"}, is essential to
ensure the reliability and effectiveness of robotic systems guided by LLMs.

Instead of relying on LLMs solely trained on language, an alternative idea is
to use the same learning architectures as LLMs (transformers), but train them
on multi-modal robotics data instead of pure text, like in the Robotics
Transformers (RT) line of work \cite{brohan2023rt}. A more popular and less
compute-intensive approach is to use similar large-scale robotics datasets and
incorporate pre-trained language and vision models with a few trained layers to
connect the components. OpenVLA
\cite{kim2024openvlaopensourcevisionlanguageactionmodel} uses this approach
with small open-source models (7 billion parameters compared to GPT-3's 175
billion.), highlighting the potential for substantial achievements with smaller
models.


Regarding humanoid robots, recent research has focused on generating human-like
motions from text descriptions, specifically through animation using simulated
human-like articulated models~\cite{ren2023insactor, motionlcm}. However, for
the problem of multi-contact planning, we are only aware of a traditional
(pre-LLMs) language processing-based approach, where an n-gram language model
is employed. The goal of this model is to learn motion as a sequence of
transitions, where each word represents a shape pose, and each sentence
represents a motion \cite{mandery2016using}.


\section{Methods}
\begin{figure*}[!thp]
    \centering
    \includegraphics[width=\textwidth]{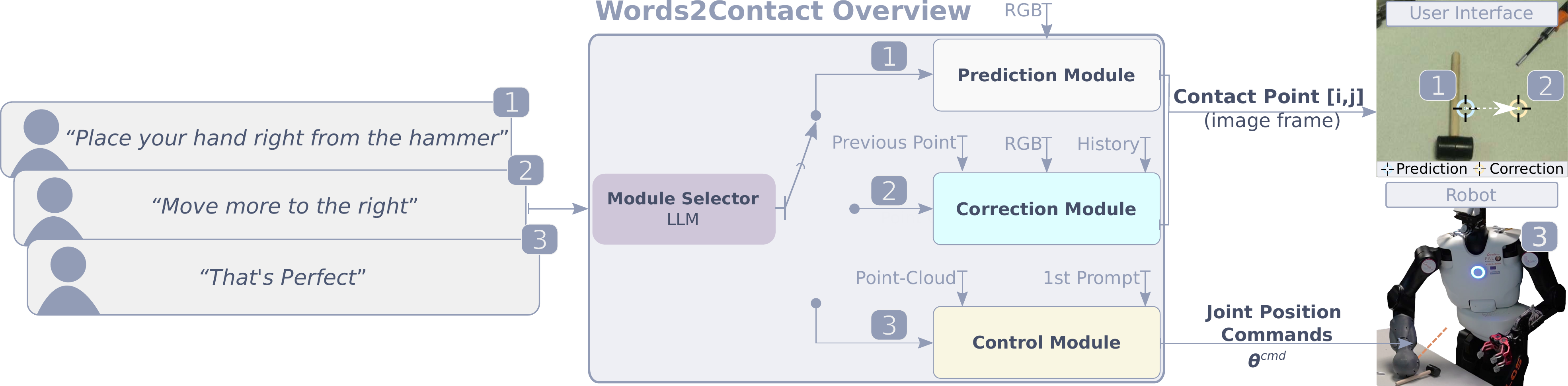}
    \caption{\textbf{Words2Contact} overview: (1) The user provides the
        first instruction. The \textbf{Module Selector}
        (Sec.~\ref{subsub:module_selector}) classifies it as ``Prediction''. The
        \textbf{Prediction Module} (Sec.~\ref{subsub:prediction_module}) integrates the
        user input and the robot's RGB data to predict a new contact point. (2) The
        user wants to adjust the predicted contact; their instruction is classified as
        ``correction''. The \textbf{Correction Module}
        (Sec.~\ref{subsub:correction_module}) adjusts the previous prediction based on
        the new user input and the RGB data. (3) The user confirms the corrected
        contact location: the instruction is classified as ``Confirmation''. The
        \textbf{Control Module} (Sec.~\ref{subsub:control_module}) uses the PointCloud,
        the initial user prompt and the desired contact location to compute the desired
        3D contact task, later executed by the SEIKO Multi-Contact whole-body
        controller.}
    \label{fig:lpvcs-pipeline}
    \vspace{-1em}
\end{figure*}

The \textbf{Words2Contact} pipeline (Fig.~\ref{fig:lpvcs-pipeline}) unfolds as
follows: visual feedback is streamed to the user, who starts by instructing the
robot to place a contact in a specified location. The initial prediction
resulting from this instruction is displayed to the user. If dissatisfied, the
user can either correct the prediction or provide a different instruction until
they confirm satisfaction with the updated predicted target. Once the contact
location is confirmed, the robot proceeds to execute the contact placement at
the specified point using the \textbf{SEIKO} controller.

To achieve this, we split the contact prediction task into three sub-modules,
each responsible for a specific sub-task: Prediction, Correction, and
Confirmation. This split is crucial for ensuring that even small models will be
able to effectively handle each stage of the pipeline.

\subsection{Prompting LLMs}

\begin{figure}[!htpb]
    \centering
    \includegraphics[width=0.95\linewidth]{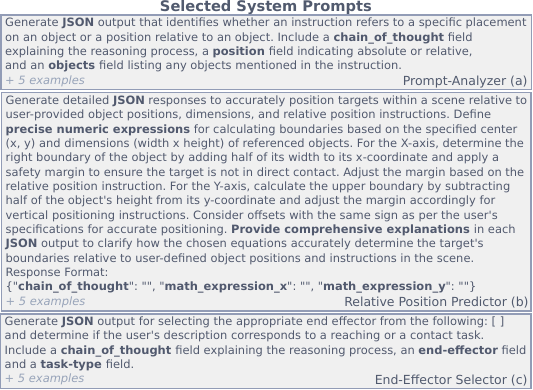}
    \caption{Some examples of system prompts that are utilized by the LLM in
        our modules. ``+5 examples'' refers to the 5 examples that are added to the
        system prompt, as part of the
        few-shot prompting technique. All the system prompts are available at:
        \url{https://hucebot.github.io/words2contact_website/}.}
    \label{fig:sys-prompts}
\end{figure}

We use a single LLM and dynamically adjust the system prompt
(Fig.~\ref{fig:sys-prompts}) at each step of the pipeline. Furthermore, outputs
from the LLM are constrained to JSON format to ensure a desired structure that
simplifies information extraction, in the open source models we enforce this
constraint with grammar-based token sampling and acceptance
\cite{willard2023efficient}, whereas for the proprietary model we follow the
documentation instructions\footnote{OpenAI Docs:
    \url{https://tinyurl.com/openaijson}}.


For readers unfamiliar with LLMs, it is important to differentiate between the
system prompt and the user prompt. The system prompt is an instruction that
guides the LLM's responses, setting the tone, context, and boundaries for the
conversation. Each module has a specific system prompt tailored to its task, as
illustrated in Fig.~\ref{fig:sys-prompts}. In contrast, the user prompt is the
natural language input or query provided by the user.

\subsection{Module Selector}
\label{subsub:module_selector}
The Module Selector (Fig.~\ref{fig:lpvcs-pipeline}) interprets the user's
natural language prompt and classifies
it into one of three categories: Prediction, Correction, or Confirmation.

This classification is achieved by combining two key techniques: few-shot
prompting~\cite{brown2020language} and logits bias\footnote{OpenAI Article:
    \url{https://tinyurl.com/openailogitbias}}. Few-shot prompting involves
providing a system prompt that describes the task that the LLM has to perform,
accompanied by examples to guide it in classifying new inputs correctly. For
this, and all the following modules, we use $5$-shot prompting, i.e., we
provide five examples. Logits bias is a technique used to adjust the output
probabilities of logits, specifically for terms such as `Prediction',
`Correction', and `Confirmation'. This adjustment aims to prioritize the
correct classification of inputs into one of these three categories.

\subsection{Prediction Module}
\begin{figure*}[!bhpt]
    \centering
    \includegraphics[width=0.95\textwidth]{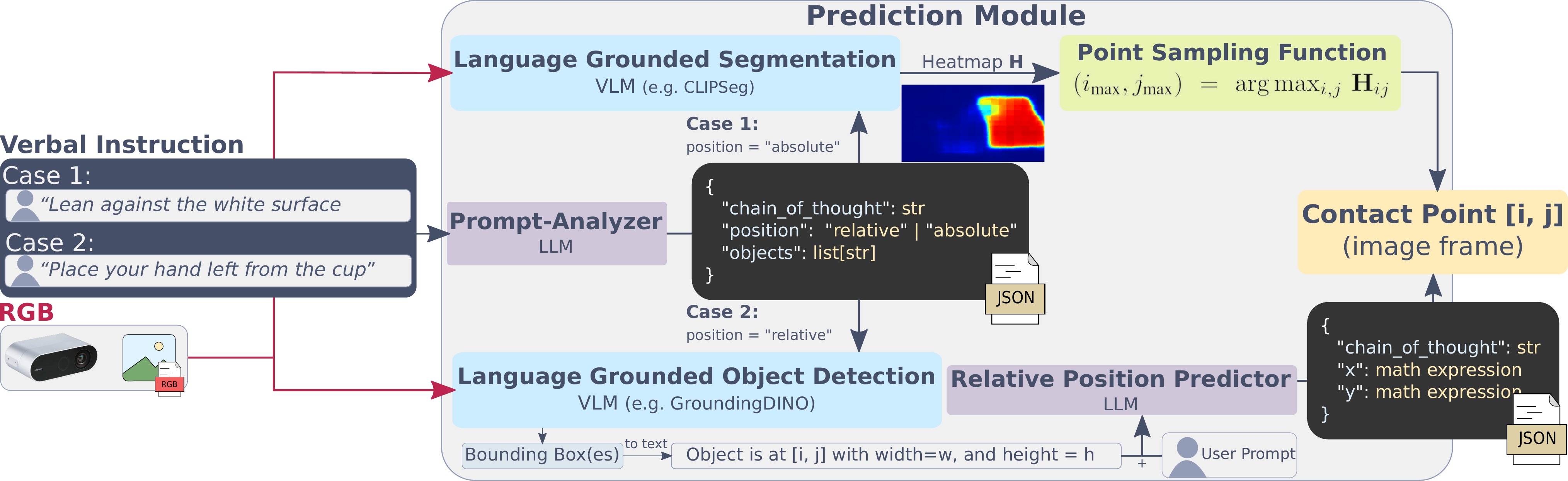}
    \caption{ The \textbf{Prediction Module}
        (Sec.~\ref{subsub:prediction_module}): the Prompt-Analyzer is an LLM that
        analyzes the user's prompt and returns a JSON file with the chain of thought,
        list of objects, and position type (absolute or relative). For absolute
        positions, a point is extracted via language-grounded segmentation. For
        relative positions, a language-grounded object detection VLM detects bounding
        box(es), used by the LLM to predict the contact point.}
    \label{fig:prediction-module}
\end{figure*}
\label{subsub:prediction_module}
To interpret the desired location implied by the user, we need to have a system
that leverages both natural language instructions and visual state feedback.
The Prediction
module (Fig.~\ref{fig:prediction-module}) combines
both Vision Language (VLMs) and a Large Language Model (LLM).
We assume that there are two cases of positions that the user might refer to:
\begin{enumerate}
    \item \textbf{Absolute Positions:} For prompts specifying a contact that is
          on an object (e.g., ``\textit{place your hand on the book}'').
    \item \textbf{Relative Positions:} For prompts where the contact is
          expressed in terms of its spatial relation to the object(s) (e.g.,
          ``\textit{left from the box}'', ``\textit{between the cup and the bowl}'').
\end{enumerate}

The \textbf{Prompt Analyzer} is responsible for (a) identifying which of the
two
scenarios the prompt is relevant to and (b) isolating the object's descriptions
mentioned in
the prompt so that they can be passed to the VLMs. For complex tasks that
involve common sense and math reasoning, chain-of-thought prompting, where the
LLM is asked to provide its thought process before reaching a conclusion, has
proven to be beneficial~\cite{wei2023chainofthought}. We combine few-shot
prompting and chain-of-thought reasoning, to achieve better results (see the
prompt on Fig.~\ref{fig:sys-prompts}-a).

In the case of \textbf{Absolute Positions} (Case 1 in
Fig.~\ref{fig:prediction-module}), we use the capability of
language-grounded segmentation models to segment images based on natural
language descriptions. This allows the system to detect the object regardless
of how
they are referenced by the user, overcoming the limitations of classic
pre-trained segmentation models that either
retrieve a mask based on a pre-trained set of labels or return a segmentation
of an image without any labeling.
\textbf{CLIPSeg}~\cite{luddecke2022image}, for example, addresses this problem
by extending a CLIP model~\cite{radford2021learning} with a transformer-based
decoder. Once we obtain the segmentation heatmap for the requested object, we
determine the coordinates of the contact point in image space using the
following metric: $[i_{\text{max}}, j_{\text{max}}] = \arg\max_{i, j} \,
    \mathbf{H}_{ij}$, where $\mathbf{H}$ is the heatmap produced by the
language-grounded segmentation model, up-scaled to the size of the original
image.	While a more sophisticated point sampling technique could be chosen to
ensure sufficient space coverage, such considerations are beyond the scope of
this work.

In the case of \textbf{Relative Positions} (Case 2 in
Fig.~\ref{fig:prediction-module}), we utilize spatial relationships
derived from the visual scene and the verbal instruction to determine the
contact location. To achieve this, following the same intuition as in the first
case, we extract the bounding box(es) using pre-trained open-set object
detection \cite{liu2023grounding}. Similarly
to grounded segmentation, these models are trained using
bounding box annotations and aim at detecting arbitrary classes with the help
of language generalization. The representation of a bounding box using natural
language is
straightforward and thus motivates our approach. For instance, in case 2 of
Fig.~\ref{fig:prediction-module}, after we receive a bounding box for the
cup, we build the following prompt: ``\textit{Cup is at [100,150] with
    width=120 and height=90. Place your hand left from the cup.}'' The system
prompt (Fig.~\ref{fig:sys-prompts}-b) contains some
basic information about the representation we are following. Additionally, we
provide a few examples to ensure that the LLM will accurately interpret the
spatial instruction accurately, and will calculate the final contact position
successfully. Furthermore, instead of directly outputting a numerical value,
the LLM outputs a mathematical expression which is then parsed and calculated
using a Python parser. This choice was made because, in preliminary
experiments, we noticed a performance increase of around $10\%$ when using this
approach instead of having the LLM perform the computation on its own.

\subsection{Correction Module}
\label{subsub:correction_module}
\begin{figure}[]
    \centering
    \includegraphics[width=\linewidth]{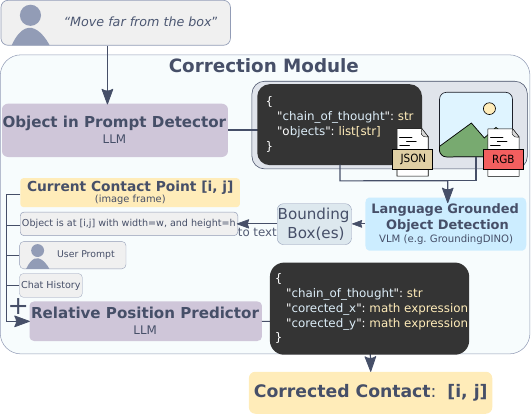}
    \caption{In the \textbf{Correction Module}
        (Sec.~\ref{subsub:correction_module}) the LLM detects object descriptions in
        the user prompt, the VLM identifies their bounding boxes, and then uses them
        along with the current target position, interaction history, and the user's
        instruction to determine a new candidate contact $[i,j]$.}
    \label{fig:correction-module}
\end{figure}

When dealing with humanoid robots and contacts, precision is of high
importance. The Correction
module (Fig.~\ref{fig:correction-module}) enhances the \textbf{Words2Contact}
pipeline
by allowing for both minor and major corrections. Similarly to the second
scenario of the Prediction Module, we detect the object(s) stated in the user
prompt, and then we retrieve their bounding boxes. The main difference is that
in the system prompt we mention that the goal now is to correct a user-given
position, and in the final user prompt, we include the current target position.
Another important point to specify is stating a correction that includes an
object (e.g., ``\textit{Move closer to the cup.}'') is optional, and we even
support prompts of the form: ``\textit{Move the target a bit to the right.}''.
Finally, to provide a more natural interaction with the correction module, we
include the conversation history, which allows the operator to make
corrections relevant to the previous ones, for example ``\textit{Move to
    the right.}'', ``\textit{Now, move twice as much as before.}''.

\subsection{Control Module}
\label{subsub:control_module}

Once the user confirms their satisfaction with the displayed contact point, we
query the LLM one final time using the end-effector selector prompt
(Fig.~\ref{fig:sys-prompts}-c), to determine the robot's end-effector (e.g.,
right or left hand) and the task type (e.g., support contact or reaching).
We extract the 3D position, in camera frame $\mathbf{p}^{cam}= [x,y,z]^{cam}$,
from the point cloud and we transform it to robot's world frame using forward
kinematics.
A spline-based Cartesian trajectory, starting from the current position
$\mathbf{x}^{EE}_t$ at time $t$, brings the end-effector EE to the desired
contact or reaching point $\mathbf{p}^{rob}$.

The Control Module utilizes SEIKO Retargeting \cite{seiko, seiko2} and
Controller \cite{rouxel2024multi} to track the effector position, smoothly
establishing new contacts by redistributing contact forces across the effector
to participate in the robot's balance.
The SEIKO pipeline takes as input the instantaneous effector target position
sampled from the spline, along with the contact force measurements from
force-torque sensors, and produces joint positions that are sent to the robot's
hardware.
SEIKO is implemented as a model-based Sequential Quadratic Programming (SQP)
optimization method to initially compute the desired whole-body configuration
based on the target effector position.
The Controller then regulates contact forces to achieve multi-contact motion
while ensuring robustness.
For safety, SEIKO enforces motion feasibility and imposes constraints on joint
position and velocity limits, quasi-static balance, and conditions to prevent
slipping and tipping.
If the pipeline receives an infeasible or unreachable position, the robot will
attempt to approach the target as closely as possible without compromising its
balance.

\begin{figure}[]
    \centering
    \includegraphics[width=\linewidth]{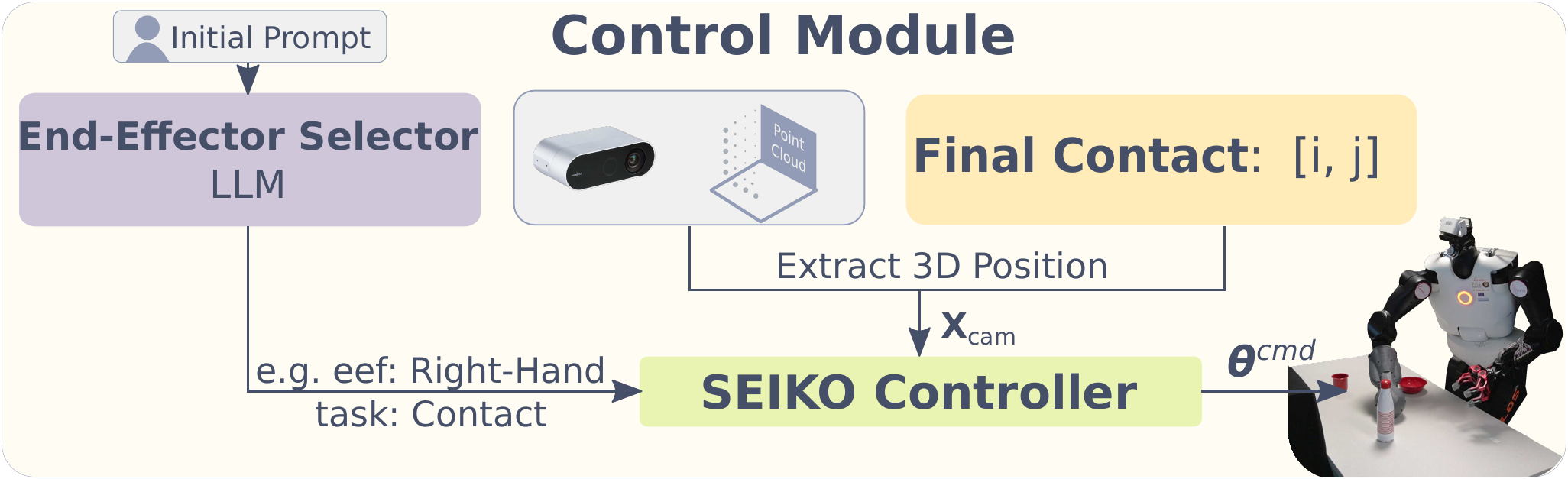}
    \caption{In the \textbf{Control Module} (Sec.~\ref{subsub:control_module}),
        the SEIKO Controller commands the robot to realize the desired task with the
        selected end-effector at the desired contact position.}
    \label{fig:control-module}
\end{figure}


\section{Experiments \& Results}


\subsection{Evaluation of contact prediction using pre-trained models}
\label{sub-section:exp_pred}
In this experiment, we benchmark several state-of-the-art pre-trained models
(VLMs and LLMs): the goal is to select the best combination for our pipeline,
evaluating the impact of the type of model and its size (i.e., the number of
parameters) on the prediction performance (mapping user inputs to pixels).

\begin{figure}[]
    \centering
    \includegraphics[width=\linewidth]{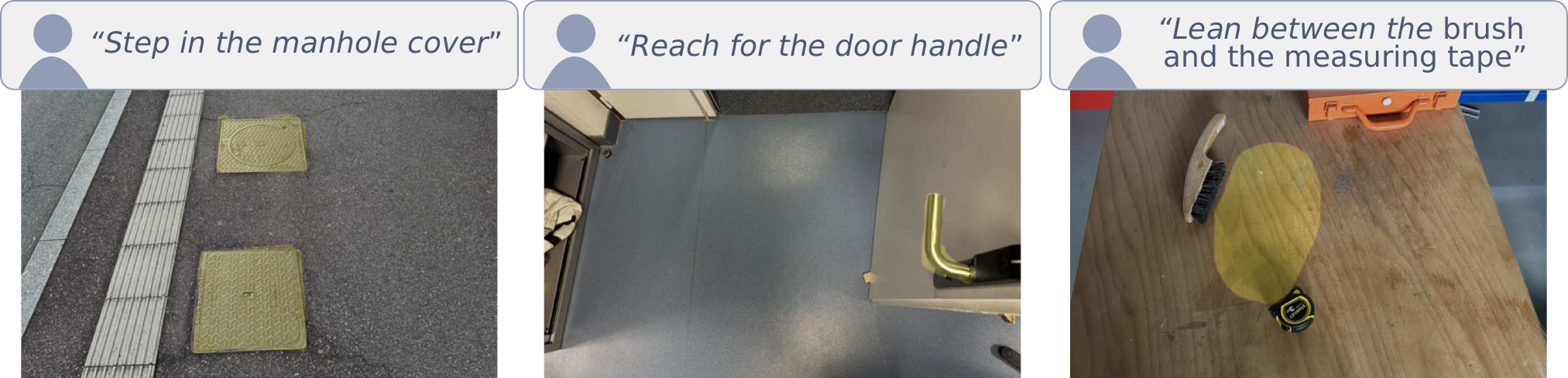}
    \caption{Selected records from our dataset. The yellow
        masks indicate acceptable contact areas for each given prompt.}
    \label{fig:dataset-demo}
\end{figure}

To this purpose, we created a new dataset\footnote{The dataset can be
    downloaded from our website:
    \url{https://hucebot.github.io/words2contact_website/}}, with $78$ tuples of
images (1280$\times$720 pixels), prompts, and manually annotated masks
corresponding to the area that satisfies the described contact. The dataset has
both indoor and outdoor images, contains different ways of requesting contacts
(e.g., {\it lean, place}), different end-effectors (e.g., {\it hand, foot}),
and an even distribution of relative and absolute positions
(Fig.~\ref{fig:dataset-demo}).

To evaluate the performance of the prediction module and assess the impact of
the LLM size, and the VLM choice on the performance of our pipeline, we
evaluate the success rate of each combination of several models: For the LLMs,
we test Calme-7b-Instruct, mixtao-7bx2-moe and GPT-3.5-turbo. While for the VLM
segmentation we chose CLIPSeg~\cite{luddecke2022image} and CLIP
Surgery~\cite{li2023clip}, and for the object detection
GroundingDINO~\cite{liu2023grounding} and Florence-2~\cite{xiao2023florence2}.
Random selection was also used to establish a baseline.

The results (Tab.~\ref{tab:success-rates}), show that all the combinations
outperform the random point sampling, and that the best combination of models
(gpt3.5+GroundingDino+ClipSeg) selects appropriate contacts in about 70\% of
the absolute cases and 50\% of the relative cases. One key finding is that
breaking down larger tasks into smaller subtasks enables smaller models to
achieve a success rate comparable to larger models, despite their significantly
reduced size. This suggests that task structuring can be a valuable strategy in
optimizing model performance, especially when computational resources are
limited.

\begin{table*}[!bhtp]
    \centering
    \caption{Success rate of each combination of foundation models.}
    \resizebox{\textwidth}{!}{%
        \begin{tabular}{lccccc}
            \toprule
            \multicolumn{3}{c}{\textbf{Combination}}                &
            \multicolumn{3}{c}{\thead{\textbf{Success Rate}}}

            \\
            \midrule
            \multirow{2}{*}{\textbf{LLM}}                           & \multirow{2}{*}{\textbf{VLM
            ObjectDetection}}                                       & \multirow{2}{*}{\textbf{VLM Segmentation}} &
            \textbf{absolute}                                       & \textbf{relative}                          & \textbf{overall}                                                                         \\
            \cmidrule{4-6}
                                                                    &                                            &                          & median [25\%, 75\%]     & median [25\%, 75\%] & median [25\%,
            75\%]                                                                                                                                                                                           \\
            \midrule
            Calme-7b-Instruct                                       & Florence-2                                 & CLIPSeg                  & $0.67$ [$0.66$, $0.68$]
                                                                    & $0.39$ [$0.39$, $0.4$]                     & $0.54$ [$0.39$, $0.66$]                                                                  \\
            Calme-7b-Instruct                                       & Florence-2                                 & CLIP Surgery             & $0.43$ [$0.42$,
            $0.45$]                                                 & $0.42$ [$0.41$, $0.43$]                    & $0.42$ [$0.42$, $0.45$]                                                                  \\
            Calme-7b-Instruct                                       & GroundingDINO                              & CLIPSeg                  & $0.71$ [$0.7$,
            $0.71$]                                                 & $0.46$ [$0.45$, $0.48$]                    & $0.59$ [$0.47$, $0.71$]                                                                  \\
            Calme-7b-Instruct                                       & GroundingDINO                              & CLIP Surgery             & $0.45$ [$0.43$,
            $0.45$]                                                 & $0.46$ [$0.43$, $0.48$]                    & $0.45$ [$0.43$, $0.47$]                                                                  \\

            \midrule
            mixtao-7bx2-moe                                         & Florence-2                                 & CLIPSeg                  & $0.66$ [$0.63$, $0.69$] &
            $0.34$ [$0.34$, $0.36$]                                 & $0.51$ [$0.34$, $0.64$]                                                                                                               \\
            mixtao-7bx2-moe                                         & Florence-2                                 & CLIP Surgery             & $0.43$ [$0.42$,
            $0.45$]                                                 & $0.36$ [$0.33$, $0.39$]                    & $0.42$ [$0.36$, $0.45$]                                                                  \\
            mixtao-7bx2-moe                                         & GroundingDINO                              & CLIPSeg                  & $0.66$ [$0.65$, $0.68$]
                                                                    & $0.41$ [$0.39$, $0.42$]                    & $0.53$ [$0.41$, $0.66$]                                                                  \\
            mixtao-7bx2-moe                                         & GroundingDINO                              & CLIP Surgery             & $0.42$ [$0.41$,
            $0.43$]                                                 & $0.45$ [$0.41$, $0.47$]                    & $0.42$ [$0.41$, $0.45$]                                                                  \\
            \midrule
            gpt-3.5                                                 & Florence-2                                 & CLIPSeg                  & $0.74$ [$0.72$, $0.74$] & $0.39$
            [$0.38$, $0.39$]                                        & $0.55$ [$0.39$, $0.73$]                                                                                                               \\
            gpt-3.5                                                 & Florence-2                                 & CLIP Surgery             & $0.42$ [$0.41$, $0.45$] &
            $0.34$ [$0.34$, $0.39$]                                 & $0.41$ [$0.36$, $0.44$]                                                                                                               \\
            \textbf{gpt-3.5}                                        & \textbf{GroundingDINO}                     & \textbf{CLIPSeg}         &
            $0.71$ [$0.68$, $0.74$]                                 & $0.5$ [$0.5$, $0.53$]                      & $\mathbf{0.61}$ [$0.51$,
            $0.7$]                                                                                                                                                                                          \\
            gpt-3.5                                                 & GroundingDINO                              & CLIP Surgery             & $0.45$ [$0.43$, $0.46$] &
            $0.53$ [$0.5$, $0.54$]                                  & $0.47$ [$0.45$, $0.51$]                                                                                                               \\
            \midrule
            \multicolumn{3}{c}{Results using Random Point Sampling} & $0.12$
            [$0.08$, $0.14$]                                        & $0.17$ [$0.11$, $0.21$]                    & $0.13$ [$0.1$, $0.18$]                                                                   \\
            \bottomrule
        \end{tabular}
        \label{tab:success-rates}
    }
    \vspace{-1em}
\end{table*}


\subsection{Pilot Study - Evaluation of the correction mechanism and usability
    of the pipeline}
\label{sub-section:exp_users}
\begin{figure}[!thbp]
    \centering
    \includegraphics[width=0.9\linewidth]{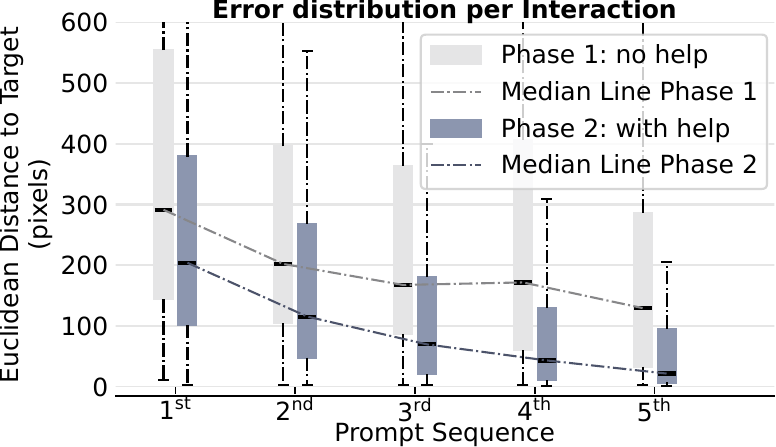}
    \caption{Results of the pilot study: all the participants were able to
        bring the predicted contact close to the target in a few iterations with our
        system, exhibiting quick learning, both with and without prompt expert
        guidance. The correction mechanism is very effective in achieving accuracy in
        contact placement, which is critical for real robot applications.}
    \label{fig:exp2_errors_boxplot}
\end{figure}
To evaluate the performance and usability of our system, we conducted a pilot
study with 11 volunteer participants ($9$ male, $2$ female, aged $26.27\pm1.8$
y.o., min $24$, max $30$). All participants had no prior experience with our
system. The pilot study was structured in two phases.

In the first phase, participants were presented with a set of $10$ images,
randomly sampled from the same dataset as in Sec.~\ref{sub-section:exp_pred}.
Each image displayed a random target
marked with a circle of 18 pixels radius; the participant's task was to tell
the system how to accurately place a point, marked with another circle of 5
pixels radius, on the designated target using a maximum of 5 prompt steps.
Each participant received minimal instructions and, notably, was not informed
about the existence of the correction module.


In the second phase, we provided the same participants with an explanation of
the system's functionalities, including two examples of prediction sentences
and two correction sentences. With this new knowledge, they instructed the
system to identify $10$ targets on $10$ different images.

We measured the distance between the predicted point and the target across
prompts, as the distance between the centers of the two circles. The results
(Fig.~\ref{fig:exp2_errors_boxplot}) demonstrate a significant improvement in
task performance in both phases: all the users quickly learned how to use the
system, and were able to bring the point close to the target with few
corrections. Their performance was, as expected, better after the prompt expert
suggestions (the median distance at the 5th prompt is 21 pixels, which is
comparable with the target circle radius).

The quick and accurate target reaching demonstrates the effectiveness of the
prediction-correction mechanism in Words2Contact for precise contact
identification, essential for real robot applications. Participants also
reported high engagement and satisfaction with the system.


\subsection{Real Robot experiment}
\label{sub-section:exp_talos}
\begin{figure*}[!htbp]
    \centering
    \includegraphics[width=0.85\textwidth]{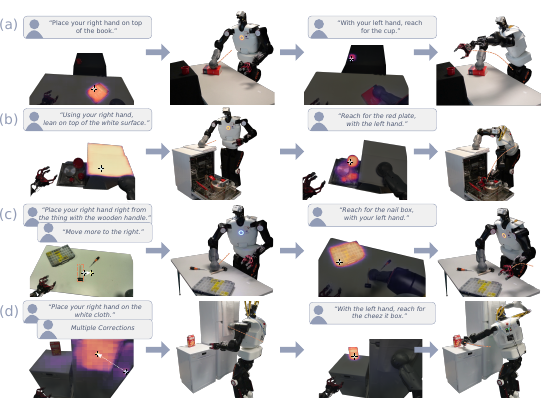}
    \caption{Talos receives instructions in natural language from an operator.
        The images show examples of sequences of reaching actions with a support
        contact in different scenarios: a) book on table, b) dishwasher bottom rack, c)
        box on table, d) cloth over the fridge. Orange dashed lines indicate the motion
        of the robot's end-effector; blue targets indicate the prediction module's
        estimated contact locations; yellow targets indicate the corrected target if
        adjustments were made. Between each execution, the operator confirmed their
        satisfaction with the predicted target location, but the confirmation steps are
        excluded for clarity. A video showing the experiments is available at
        \url{https://hucebot.github.io/words2contact_website/}.}
    \label{fig:robot_exp_1}
    \vspace{-2em}
\end{figure*}

We evaluated \textbf{Words2Contact} with the Talos humanoid robot
\cite{stasse2017talos} in four distinct whole-body reaching settings
(Fig.~\ref{fig:robot_exp_1}), with and without corrections, with the following
user prompts \circled{p}:
\begin{itemize}
    \item[(a)] \circled{1} ``{\it Place your right hand on top of the book}''
        \circled{2} ``{\it with your left hand, reach for the cup}''.
    \item[(b)] \circled{1} ``{\it Using your right hand, lean on top of the white
        surface}'' \circled{2} ``{\it reach for the red plate, with the left hand}''.
    \item[(c)] \circled{1} ``{\it Place your right hand right from the thing with
        the wooden handle}'' (this is a mallet, but the user might not know the name);
        operator's correction to avoid a collision: \circled{2} ``{\it Move more to the
        right}''; \circled{3} ``{\it Reach for the nail box, with your left hand}''.
    \item[(d)] \circled{1} ``{\it Place your right hand on the white cloth}''; 6
        corrections \circled{2}-\circled{7} guide the target, as the setting lacks
        distinct objects for relative positioning; \circled{8} ``{\it with the left
        hand, reach the cheez it box}''.
\end{itemize}

We highlight that the operator completed all the tasks
using 
paraphrases to describe objects (``{\it thing with a wooden handle}'') and rare
names (``{\it cheez it box}''). The robot always performed the task
successfully.

\medskip

\textbf{Video/Code/Dataset: }
The video of the robot experiments, the dataset of
Sec.~\ref{sub-section:exp_pred} and the software to reproduce
\textbf{Words2Contact} are available at
\url{https://hucebot.github.io/words2contact_website/}.



\section{Conclusions \& Future Work}

\textbf{Words2Contact} is especially useful in scenarios where users cannot
directly control the robot, such as in human-robot collaboration settings and
teleoperation. For instance, in a collaborative task, the robot could
autonomously choose and adjust its hand placement on a shared work surface
while the human partner focuses on a different aspect of the task. While the
system adapts well in these contexts, the confirmation of the final contact
point remains an open question. In this work, we mainly focused on
language-based teleoperation, where the operator is remote from the scene and
can confirm the location of the final contact point through visual feedback;
however an equivalent confirmation system will also be necessary in human-robot
collaboration. In our pilot study, users quickly adapted to the system, and
real-world experiments demonstrated that Words2Contact delivers satisfactory
contact placements, even in challenging environments, through its iterative
correction mechanism. Based on these findings, we expect similar performance in
human-robot collaboration, where a reliable confirmation system could ensure
precise contact point adjustments during shared tasks.

For future work, we aim to reduce user reliance on visual feedback by enabling
online corrections and scene-grounded trajectory generation for the
end-effectors~\cite{shi2024yell,cui2023onlinelanguagecorrections,kwon2024language}.
Additionally, improving the system's performance will involve ensuring motion
feasibility through real-time evaluation of inverse kinematics and surface
safety using VLMs. For example, predicted contact points will be constrained to
areas that are both a) reachable and b) safe, such as avoiding fragile surfaces
like glass.

Finally, Words2Contact opens up opportunities for shared or full autonomy by
capturing human-robot interaction data during teleoperation to build a robust
dataset. This data could be used to fine-tune LLMs or train imitation learning
approaches for shared autonomy~\cite{rouxel2024flowmatchingimitationlearning},
allowing the system to autonomously manage routine tasks while requiring human
intervention only for more complex decisions.





\raggedright
\printbibliography


\end{document}